# Enhanced Teaching-Learning-based Optimization for 3D Path Planning of Multicopter UAVs

Van Truong Hoang[1], Manh Duong Phung[2]

[1] Naval Academy, Viet Nam
[2] University of Technology Sydney (UTS), Australia

Vantruong.Hoang@alumni.uts.edu.au, Manhduong.phung@uts.edu.au

**Abstract.** This paper introduces a new path planning algorithm for unmanned aerial vehicles (UAVs) based on the teaching-learning-based optimization (TLBO) technique. We first define an objective function that incorporates requirements on the path length and constraints on the movement and safe operation of UAVs to convert the path planning into an optimization problem. The optimization algorithm named Multi-subject TLBO is then proposed to minimize the formulated objective function. The algorithm is developed based on TLBO but enhanced with new operations including mutation, elite selection and multi-subject training to improve the solution quality and speed up the convergence rate. Comparison with state-of-the-art algorithms and experiments with real UAVs have been conducted to evaluate the performance of the proposed algorithm. The results confirm its validity and effectiveness in generating optimal, collision-free and flyable paths for UAVs in complex operating environments.

## 1 Introduction

Unmanned aerial vehicles (UAVs) have potential use in many areas of real-world applications due to their cost-effective, traffic-free, and flexible operation. Especially, they have been receiving great interest in civil applications ranging from data collection to defect and damage detection [1]. The key to the success of those applications involves path planning algorithms that generate optimal paths for UAVs to follow during their operation [2-4].

In the literature, UAV path planning has been addressed with different approaches such as the A-star probabilistic road map, fast marching square and visibility binary tree algorithms [5-8]. Besides that, nature-inspired algorithms such as the Genetic Algorithm (GA), Ant Colony Optimization (ACO), and Particle Swarm Optimization (PSO) have been successfully applied to solve challenging path planning problems that require not only feasible but also optimal paths. Depending on its advantages and limitations, a certain method can be preferable for a specific task than the others. GA has a great capacity to solve various path planning problems [9][10], but it requires heavy computation and has a premature convergence issues [4]. ACO can provide short and collision-free paths [11], but its convergence speed is rather slow [4]. PSO on the other hand is known to have good convergence speed with many variants making it suitable for a number of path planning problems [12-13]. It however also has the issue with premature convergence and in some situations may result in local optimum rather than global optimum solutions [4],[14].

Recently, a new nature-inspired optimization method named teaching-learning-based optimization (TLBO) has been introduced to solve complex optimization problems with the capacity to yield quality solutions with a high convergence rate [15],[16]. Its structure significantly decreases difficulties in parameter tuning and thus results in better execution time. In addition, the simplified TLBO (STLBO) [17] was later introduced maintaining advantages of the original TLBO but using a simpler structure to enhance performance in solving complex. Nevertheless, all of the teaching and learning activities in STLBO are constrained to one subject that may limit its capability to find the global optimum solution in problems that involve high dimension decision variables such as 3D UAV path planning.

In this study, we propose a new approach, namely the multi-subject TLBO (MS-TLBO) to deal with the path planning problem for UAVs. It is capable of generating not only flyable but also short and collision-free paths for UAVs to carry out their assigned tasks. The process starts from a 3D model of the target area, including its surrounding environment obtained through satellite maps. The MS-TLBO is then conducted to produce reference paths for the UAV that satisfy the safe and dynamic requirements. This optimization process is carried out based on a multi-constraint cost function allowing to increase the capability of obstacle avoidance and task performance. Extensive simulations, comparisons, and experiments have been conducted to demonstrate the feasibility and effectiveness of the proposed algorithm.

The rest of this paper is constructed as follows. The path planning problem is presented in Section 2. The design of the path planning algorithm using MS-TLBO is described in Section 3. Section 4 presents simulation and experimental results. A conclusion and suggestion for future work is outlined in Section 5.

## 2 UAV path planning problem

In civil infrastructure inspection applications, the UAV is required to fly along a certain path that allows it to collect the necessary data to fulfil the task. The path is typically optimal in a certain criterion, such as minimizing the flight distance or maximizing the coverage. Furthermore, factors relating to UAV safety and constraints on its maneuverability should also be considered. Those criteria can be incorporated in an objective function of the form:

$$J(T_i) = \sum_{m=1}^{M} \beta_m J_m(T_i), \tag{1}$$

where $T_i$ is the $i$th generated path for the UAV, $J_m(T_i)$ is the cost associated to a certain criterion of the $i$th path, $\beta_m$ is the weight chosen for each cost element. In this work, we focus on three main cost elements ($M$ = $3$) including the path length, collision avoidance, and operating altitude.

To find cost $J_1$ associated to the path length, the desired path $T_i$ is divided into $L$ segments, each segment is assumed to be straight, and $L$ is represented by its end-points $P_l = \{z_l, y_l, z_l\}, l = 0 \cdots L$. Thus, $J_1$ can be computed as:

$$J_1(T_i) = \sum_{l=1}^{L} \left\| P_{l+1} - P_l \right\|, \tag{2}$$

where $\|\cdot\|$ represents the Euclidean norm.

To find cost $J_2$ relating to collision avoidance, we base on the distance between the path segments and the obstacles presented in the operating area of the UAV. Specifically, denote $K$ as the total number of obstacles, cost $J_2$ computed for $L$ path segments and $K$ obstacles at each iteration can be expressed as:

$$J_2(T_i) = \frac{1}{LK} \sum_{l=1}^{L} \sum_{k=1}^{K} \max\left(1 - \frac{d_{l,k}}{r_{l,k}^S}, 0\right), \tag{3}$$

where $r_{l,k}^S$ is the safe range from the UAV to the $k$th obstacle, and $d_{l,k}$ is the distance from obstacle $k$ to the center of the $l$th segment.

Lastly, cost $J_3$ relating to altitude constraints that restrict the UAV to travel within a predefined height range can be represented via the minimum and maximum height values, $z_{\min}$ and $z_{\max}$, as:

$$J_3(T_i) = \sum_{l=1}^{L} \delta_l$$

$$\delta_l = \begin{cases} z_l^M - z_{\max}, & \text{if } z_l^M > z_{\max} \\ 0, & \text{if } z_{\min} \ge z_l^M \ge z_{\max} \\ z_{\min} - z_l^M, & \text{if } z_{\min} > z_l^M > 0 \\ \infty, & \text{if } 0 \ge z_l^M \end{cases} \tag{4}$$

where $z_l^M$ is the height of segment $l$. By formulating the cost elements as in (2) – (4) and the overall cost function as in (1), the path planning problem is converted to an optimization where the objective is to find a path $T_i$ that minimizes the cost $J$.

## 3 Path planning using enhanced teaching-learning-based optimization

Given the path planning problem defined in (1), we propose in this section our development of MS-TLBO that is capable of finding the optimal solution to minimize the cost function with details as follows.

### 3.1 Teaching-learning-based optimization (TLBO)

Teaching-learning-based optimization [18] is a bio-inspired algorithm designed on the basis of teaching and learning phenomena in a classroom in which the optimization variables are modelled via students' knowledge. The algorithm simulates two key modes of learning, i.e., through teacher (known as Teacher phase) and with learners (known as Learner phase). In the Teacher phase, the teacher is considered as the most knowledgeable person to inspire the entire class and improve students' knowledge. In the Learner phase, students further gain their knowledge by learning among themselves. The process of TLBO algorithm is explained in detail as below.

*1) Teacher phase:* In this phase, the teacher inspires students to reach his knowledge level. The students, on the other hand, try to study from the teacher to improve their knowledge. Thus, students' achievements are built up by two main factors, the teacher's knowledge, and their own quality. The average grade of the entire class is calculated based on the capability of students.

Let $N$ be the number of class members, $M$ is the mean grade of the class, and $T$ is the best solution representing the teacher's knowledge. In the first stage, the teacher attempts to train $M$ to reach his level $T$. This approach is mathematically formulated as below:

$$A_i^{new} = A_i^{old} + r_1 \left( T - T_F M \right) \tag{5}$$

where $A_i^{new}, i = 1, 2, \cdots, L$ represents a newly generated solution, $r_1$ is a random number in the range of [0, 1], $T_F$ is randomly set to either 1 or 2 with equal probability, $T_F = round\left[1 + rand\left(0,1\right)\left\{2-1\right\}\right]$. The new value of $A_i^{new}$ is incorporated to the next iteration if and only if it is better than its previous one, $A_i^{old}$.

*2) Learner phase:* In the learner phase, students will try to gain their knowledge by studying from each other. A learner interacts randomly with the others in different studying actions, for example, discussions, presentations, or group studies. In this phase, two random students $m$ and $n, \forall m \neq n \neq i$ are selected within the class to update their knowledge by the following equation:

$$\begin{cases} A_i^{new} = A_i^{old} + r_i \left( A_m - A_n \right) & if\ P_m < P_n \\ A_i^{new} = A_i^{old} + r_i \left( A_n - A_m \right) & otherwise \end{cases} \tag{6}$$

Similar to the teacher phase, $A_i^{new}$ will be added to the next iteration only if its value is better than its previous one.

### 3.2 Multi-subject TLBO for UAV path planning

Originally, TLBO was developed for a class with a single subject, which showed effectiveness and efficiency in comparison to other optimization methods [19]. However, for complex optimization problems such as path planning, the algorithm may fail to obtain the global optimum due to its limitation in handling high-dimensional variables. To overcome that limitation, we propose in this study a multi-subject approach in combination with the use of local search and an elite strategy as in [17] for TLBO. Specifically, a mutation operation is introduced to support the last worst learner as follows:

$$A_{1,k}^{new} = \begin{cases} A_{1,k} + mu, & \text{if rand}\left[0,1\right] < \mu \\ A_{1,k}, & \text{otherwise} \end{cases} \tag{7}$$

where $A_{1,k}^{new}$ is a new solution found around the current best teacher, $k \in \left\{1, 2, \cdots, D\right\}$ denotes the $k$th dimension of the decision variable with $D$ dimensions, $\text{rand}\left[0,1\right]$ is a uniformly distributed real value in the range of [0,1], $\mu$ is the mutation probability that helps to improve the local search performance, and $mu$ is a random mutation variable used in the local search operation. Let $FES$ and $max_FES$ be respectively the current and maximum number of objective function evaluations, the mutation probability $\mu$ is computed as:

$$\mu = 1 - FES / Max_FES .$$

On the other hand, $mu$ is computed by employing the chaotic sequence to enrich the mutation behavior:

$$mu = 2 \times X_n - 1, \tag{8}$$

where $X_n$ is the value of *n*th chaotic iteration and can be calculated as:

$$X_{n+1} = 4.0 \times X_n \times (1 - X_n), \tag{9}$$

where $X_0$ is taken from [0,1].

Through the mutation variable, the local search attempts to find a better solution around the current best teacher. If the new solution is better than the worst one, the elite strategy will be used to replace the worst learner with the new solution which can be formulated as follows:

$$S_w = \begin{cases} T_{1,k}^{new}, \text{if} f\left(T_{1,k}^{new}\right) < f\left(S_w\right) \\ S_w, \quad \text{otherwise} \end{cases} \tag{10}$$

where $S_w$ represents the worst learner in the current class and $f(\cdot)$ represents the objective function value of the corresponding solution.

After conducting the mutation and elite selection strategies, we further improve the quality of solutions by using a multi-subject approach as in [19]. The teacher phase in (5) is first replaced by the STLBO with elite strategy (7)-(10). In the learner phase, students will try to gain their knowledge by learning from each other. A learner interacts randomly with the others in different studying actions within *m* subjects. Two random solutions $i, k, \forall i \neq k$ are then selected and the solution is updated as:

$$\begin{cases} A_{i,j}^{new} = A_{i,j}^{old} + r_{ij}\left(A_{i,j}^{old} - A_{k,j}^{old}\right) & \text{if } P_m < P_n \\ A_{i,j}^{new} = A_{i,j}^{old} + r_{ij}\left(A_{k,j}^{old} - A_{i,j}^{old}\right) & \text{otherwise} \end{cases} \tag{11}$$

### 3.3 Algorithm Implementation

To implement the proposed MS-TLBO, initial parameters, such as coordinates of the working area, location of obstacles, and the start and destination locations of the UAV, is first determined based on data loaded from satellite maps. The MS-TLBO algorithm is then carried out to find an optimal path as in the pseudo-code shown below:

```
/* Preparation:                                                              */
Determine the working area, UB (Upper boundary), LB (Lower Boundary) using a satellite map;
Select the start and destination positions;
Identify all obstacles;
Save all the above data in an initialization file;
Compute constraints of the actual UAV;
/* Initialization:                                                           */
Load the initialization file and then the constraint file to global memory;
Initialize algorithm parameters, i.e., number_of_iteration, Best_Costs, number_of_population,
number_of_subject, number_of_decision_variable, Max_FES, and violation_cost;
FES = 0
foreach i < number_of_population do
    foreach k < number_of_decision_variable do
        P(i,k) = rand x [UB_k - LB_k] + LB_k
    end
end
Evaluate the objective function values of P;
FES = FES + number_of_population
X = rand                                    /* initialization of chaos map */
/* Path Planning:                                                            */
/* Teacher phase redefinition:                                               */
While FES < Max_FES do
    Update the chaos map value;                             /* using (9)   */
    foreach i < number_of_decision_variable do
        if rand < 1 – FES/Max_FES then
            Update the redefined teacher A_{1,k}^{new}      /* using (7)   */
        end
    end
    Evaluate the objective value of new individual;
    FES = FES + 1
    if A_{1,k}^{new} > S_w then
        Update the worst learner S_w by the new & better solution;   /* using (10) */
```

```
    end
    /*  Learner phase:                                                   */
    foreach i < number_of_population do
        foreach i < number_of_subject do
            Select other two learners randomly: m ≠ n
            if learner m is better than learner n then
             |  Update the learners' knowledge;        /*  using 1st equation in (12)   */
            else
             |  Update the learners' knowledge;        /*  using 2nd equation in (12)   */
            end
            Summary of all solutions\;
        end
        Evaluate the objective function values of A;
        FES = FES + number_of_population;
    end
    Check Violation cost;                                  /*  using (3)    */
    Evaluate each path based on the Best_Costs and violation_cost;
    Update each personal_best and global_best positions;
    Update A and Violation cost;          /*  using (11), (12) and (3), respectively   */
end
The optimized path T* is achieved.
```

**Fig. 1.** *Pseudo code of the proposed MS-TLBO algorithm*

# 4 Experiments

To evaluate the performance of the proposed algorithm for UAV path planning, we have conducted a number of experiments with details as follows.

## 4.1 Experimental setup

The UAV used in this work is a 3DR Solo drone retrofitted with additional devices as shown in Fig. 2. It has three onboard processors; one is ARM Cortex A9 for data processing and the others are Cortex M4 for flight control.

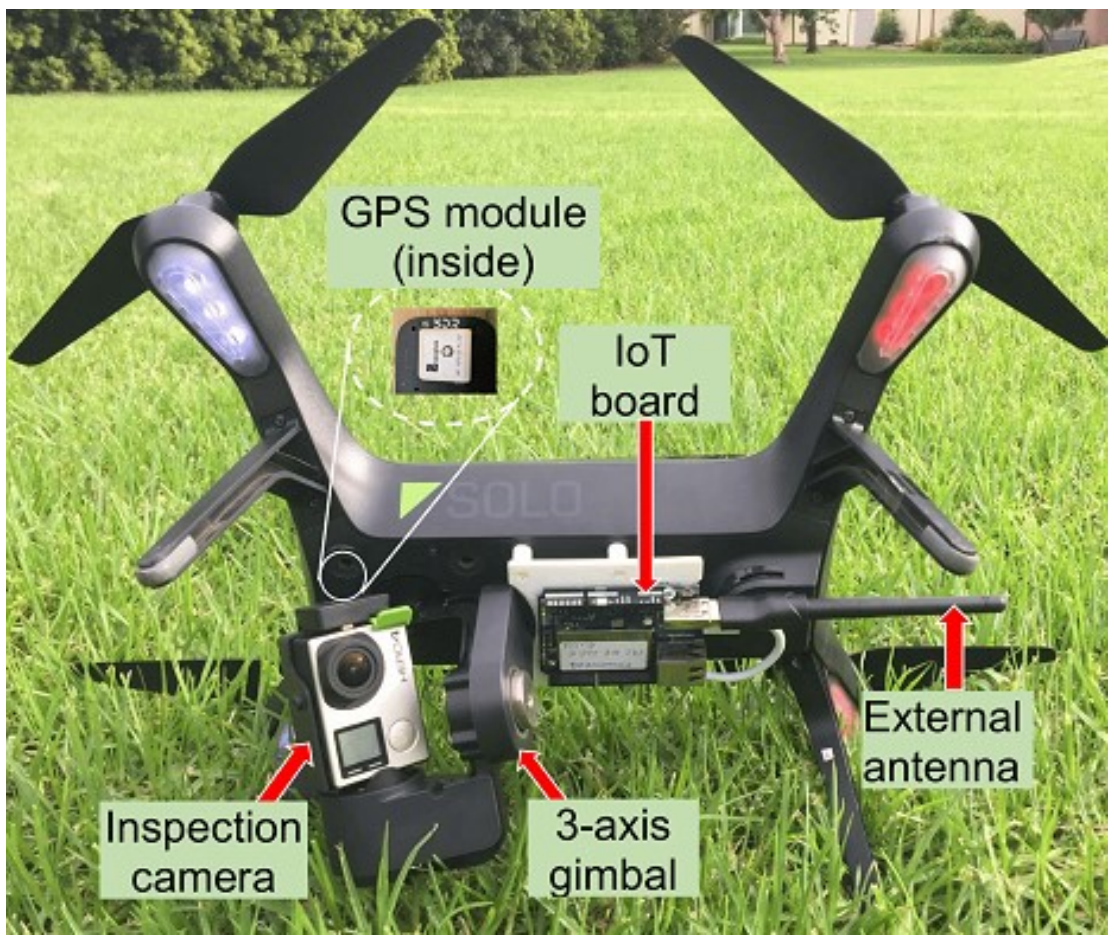

**Fig. 2.** *The retrofitted 3DR Solo quadcopter*

The retrofitted devices include a differential GPS module, an external antenna, an IoT board, and an inspection camera. The differential GPS module is a u-blox NEO-M8P-2-10 Real-time kinematic (RTK) compatible GPS receiver to increase the positioning accuracy. The inspection camera is a Hero 4 camera with 12-megapixel resolution attached to a three-axis 3D gimbal for data acquisition. The Internet-of-Things (IoT) board named Arduino-Yun is attached to the accessory port of the drone and interfaced with the embedded Linux operating system via USB protocol. Its communication range is enhanced by an extended antenna. A TP-Link wireless N 4G LTE gateway router is employed as the main access point for data communication between the drone and ground control station.

Fig. 3 shows the testing area with data acquired from a real satellite map. In this selected area, the drone is operating within the space of 90.5 m x 50.5 m x 20 m corresponding to the latitude, longitude, and altitude coordinates of {12.184706, 109.162841, 0} and {12.184253, 109.163680, 20}. The start position of the drone is

chosen at $P_s$ = {12.184333, 109.163583} and its final position is set to $P_f$ = {12.184594, 109.163040}. In the testing area, a number of obstacles, mainly due to trees, with different radii are identified and represented by pole shapes as shown in Fig. 3. Those obstacles will be incorporated as inputs to the path planning algorithm.

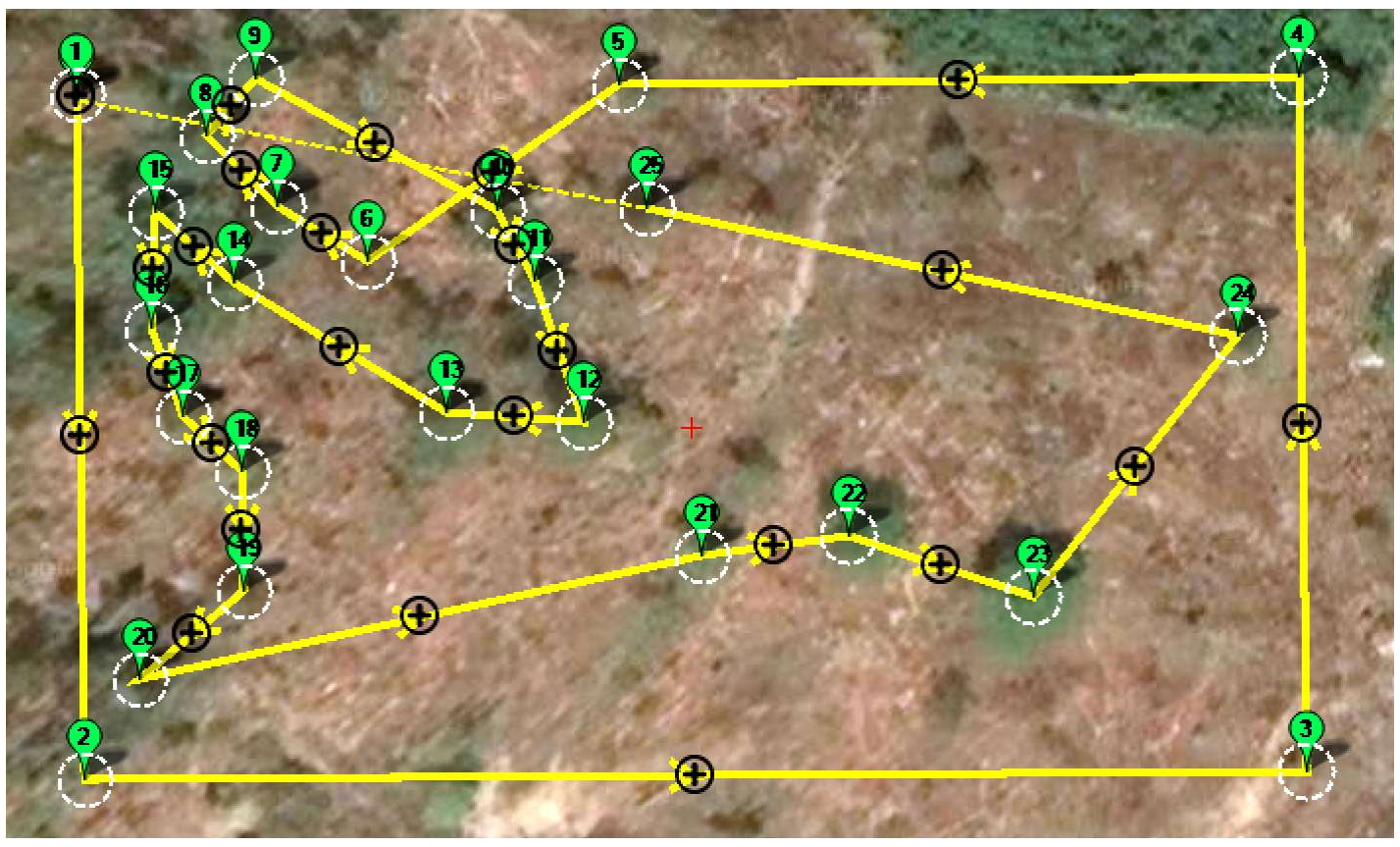

***Fig. 3.*** *The experimental area loaded from a satellite map with obstacles indicated by the black pole shape.*

In our experiment, the generated path is exported as waypoints consisting of the coordinates that the drone needs to fly through. Those waypoints are then uploaded to the drone via a ground control station software such as Mission Planner. The built-in flight controller finally controls the drone to fly along with those waypoints. In our evaluation, the number of waypoints chosen is 10 resulting in 30-dimension variables to be optimized. The terminal criterion (*Max_FES*) is set to 150,000.

### 4.2 Experimental Results

The path planning results are first compared with state-of-the-art algorithms including TLBO [15], Genetic Algorithm (GA) [9], and Ant Colony Optimization (ACO) [11]. Fig. 4 shows the cost values over iterations, wherein the MS-TLBO algorithm exhibits a better and more stable conversion. The results are confirmed as recorded in Table 1 showing the cost values and convergence iterations.

**Table 1.** Performance comparison among GA, ACO, TLBO and MS-TLBO

| Algorithm | Initial cost | Min cost | Iterations |
|---|---|---|---|
| GA | 95.912 | 73.40 | 69 |
| ACO | 81.865 | 71.80 | 107 |
| TLBO | 95.912 | 70.20 | 111 |
| MS-TLBO | 81.865 | 70.00 | 54 |

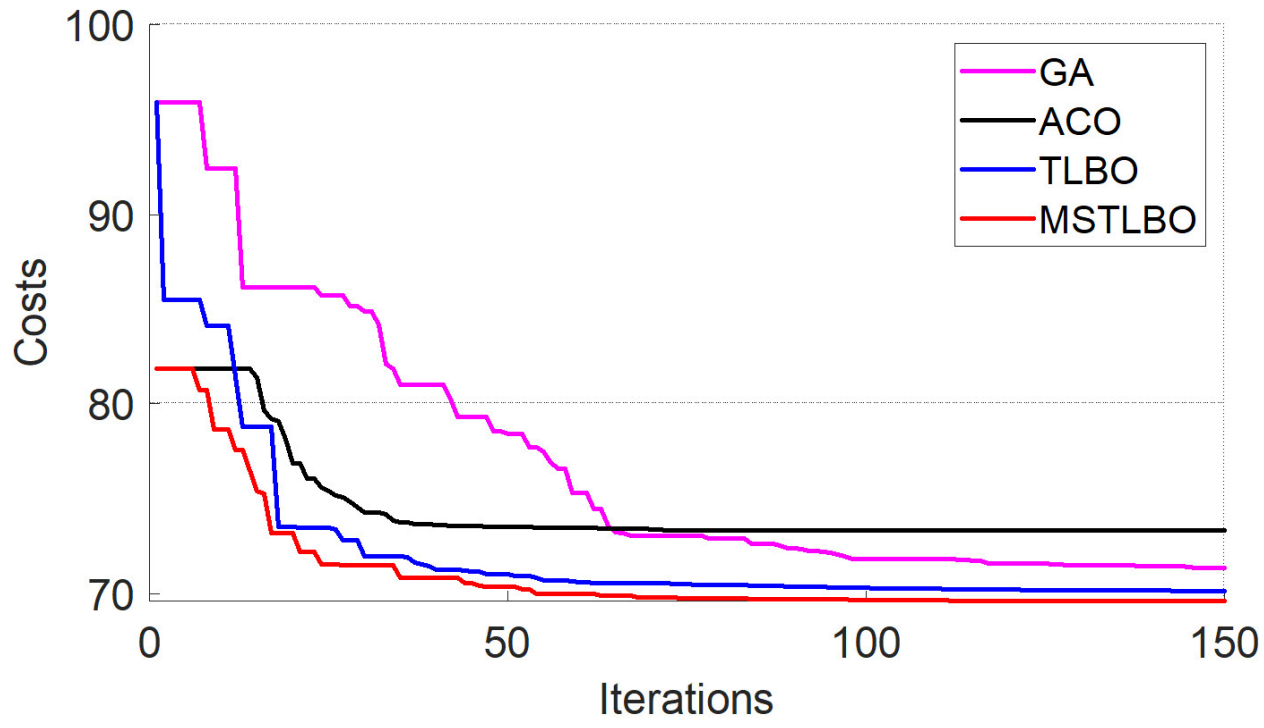

***Fig. 4.*** *Convergence rate of GA, ACO, TLBO and MS-TLBO*

The 3D path generated for the drone is shown in Fig. 5a where it can be seen that the path does not contact any obstacles. This can be further verified in Fig. 5b that shows the top view. To verify the feasibility of the planned path, its waypoints have been uploaded to the drone for real flight tests. The real trajectory of the drone is shown in Fig. 6 which overlaps the planned one. It also guides the UAV to avoid obstacles presented in the environment. The results, therefore, confirm not only the optimality of the generated path but also its validity in practical implementation.

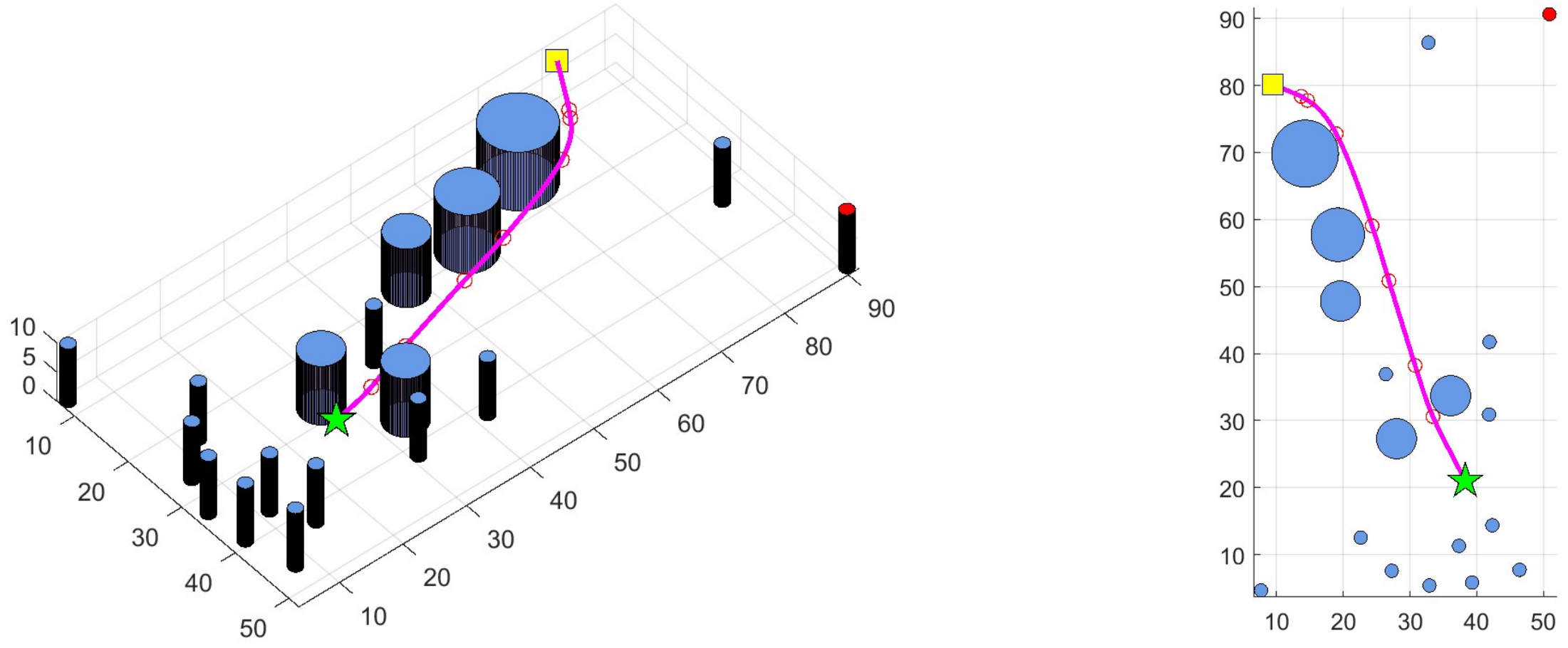

***Fig. 5.*** *The 3D path generated by MS-TLBO (a) and its top view (b)*

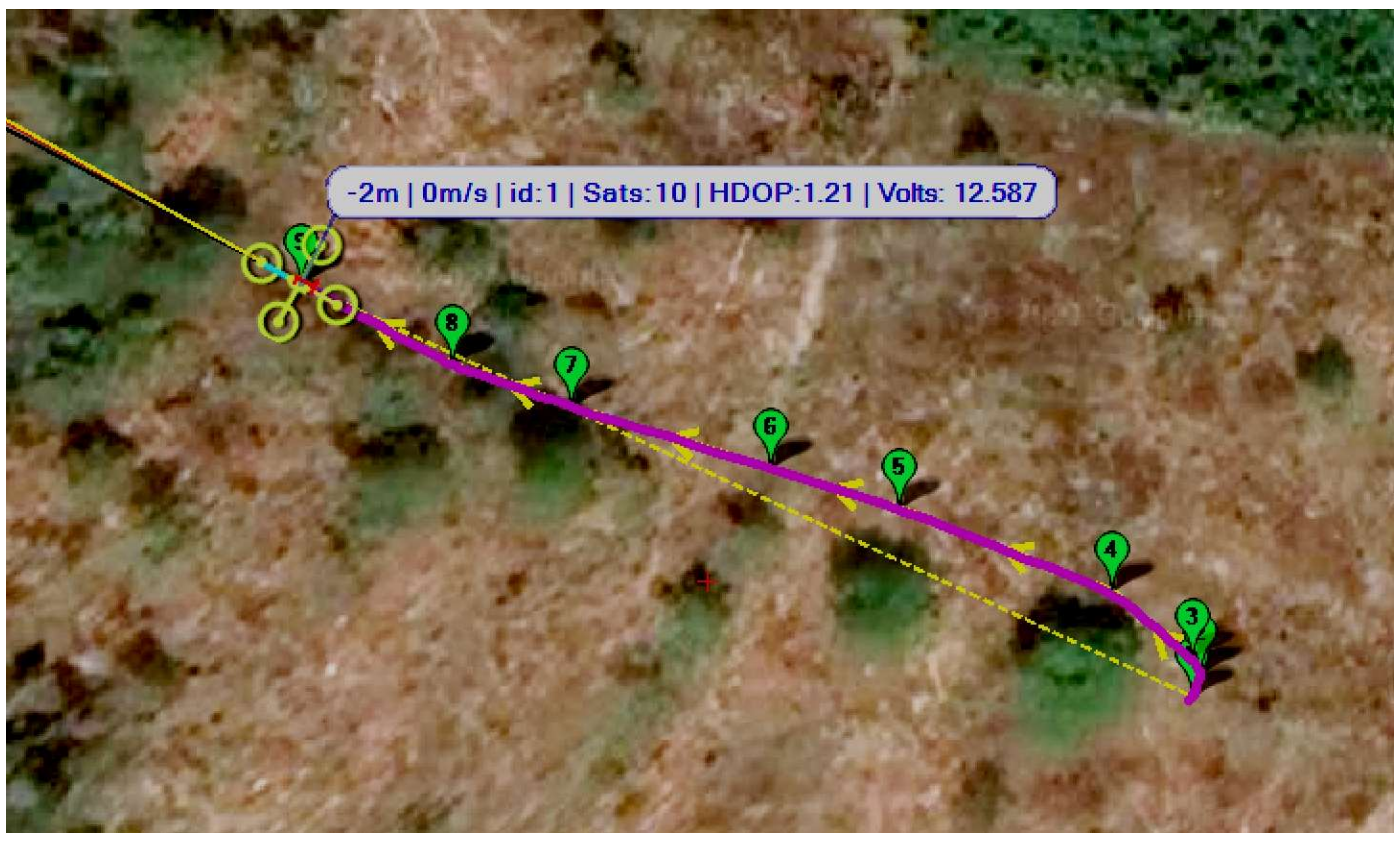

***Fig. 6.*** *The real flight path (solid purple line) overlapping the planned path (solid yellow path with arrows)*

## 3 Conclusion

In this paper, we have introduced a path planning algorithm named MS-TLBO for UAVs to conduct surface inspection tasks. The core of the algorithm is the introduction of mutation, elite selection and multi-subject operations to augment the TLBO for complex path planning tasks. When combining with a cost function that takes into account the constraints on path length, collision avoidance, and flight altitude, the algorithm can generate paths that are not only optimal in length but also feasible in movement range and safe in operation. Experimental results confirm that the proposed MS-TLBO algorithm performs better than some state-of-the-art algorithms in both optimality and convergence speed. The generated paths are also feasible for real UAV operations in a number of validation experiments. The obtained results not only confirm the validity of our approach but also suggest a possibility of extending the work toward a generic architecture for UAV path planning in complex environments.

## References

[1] A. Ellenberg, A. Kontsos, F. Moon, and I. Bartoli., Bridge deck delamination identification from unmanned aerial vehicle infrared imagery, Automation in Construction 72 (2016) 155–165 doi: 10.1016/j.autcon.2016.08.024

[2] V. T. Hoang, M. D. Phung, T. H. Dinh, and Q. P. Ha., System architecture for real-time surface inspection using multiple uavs, IEEE Systems Journal 14.2 (2019) 2925-2936 doi: 10.1109/JSYST.2019.2922290

[3] M. D. Phung, C. H. Quach, T. H. Dinh, and Q. Ha., Enhanced discrete particle swarm optimization path planning for uav vision-based surface inspection, Automation in Construction 81 (2017) 25–33 doi: 10.1016/j.autcon.2017.04.013

[4] T. T. Mac, C. Copot, D. T. Tran, and R. De Keyser., Heuristic approaches in robot path planning: A survey, Robotics and Autonomous Systems 86 (2016) 13–28.

[5] Mandloi, D., Arya, R. & Verma, A.K., Unmanned aerial vehicle path planning based on A* algorithm and its variants in 3d environment. Int J Syst Assur Eng Manag 12, (2021) 990–1000 doi: 10.1007/s13198-021-01186-9
[6] F. Yan, Y.-S. Liu, and J.-Z. Xiao., Path planning in complex 3d environments using a probabilistic roadmap method, Int. J. Automation and Computing 10. 6 (2013) 525–533
[7] C. Arismendi, D. A´ lvarez, S. Garrido, and L. Moreno., Nonholonomic motion planning using the fast marching square method, International Journal of Advanced Robotic Systems 12.5 (2015) 56-71 doi: 10.5772/60129
[8] A. T. Rashid, A. A. Ali, M. Frasca, and L. Fortuna., Path planning with obstacle avoidance based on visibility binary tree algorithm, Robotics and Autonomous Systems 61. 12 (2013) 1440–1449 doi: 10.1016/j.robot.2013.07.010
[9] V. Roberge, M. Tarbouchi, and G. Labont., Fast genetic algorithm path planner for fixed-wing military uav using gpu, IEEE Transactions on Aerospace and Electronic Systems 54.5 (2018) 2105–2117 doi: 10.1109/TAES.2018.2807558
[10] Y. V. Pehlivanoglu., A new vibrational genetic algorithm enhanced with a voronoi diagram for path planning of autonomous UAV, Aerospace Science and Technology 16.1 (2012) 47–55 doi: 10.1016/j.ast.2011.02.006
[11] B. Englot and F. Hover., Multi-goal feasible path planning using ant colony optimization, in 2011 IEEE International Conference on Robotics and Automation (2011) 2255–2260
[12] Y. Fu, M. Ding, C. Zhou, and H. Hu., Route planning for unmanned aerial vehicle (uav) on the sea using hybrid differential evolution and quantum-behaved particle swarm optimization, IEEE Transactions on Systems, Man, and Cybernetics: Systems 43.6 (2013) 1451– 1465 doi: 10.1109/TSMC.2013.2248146
[13] M. D. Phung and Q. P. Ha, Safety-enhanced UAV path planning with spherical vector-based particle swarm optimization, Applied Soft Computing 107 (2021) 107376 doi: 10.1016/j.asoc.2021.107376
[14] H. Duan, P. Li, Y. Shi, X. Zhang, and C. Sun., Interactive learning environment for bio-inspired optimization algorithms for uav path planning, IEEE Trans. Educat., 58.4 (2015) 276–281 doi: 10.1109/TE.2015.2402196
[15] G. Yu, H. Song, and J. Gao., Unmanned aerial vehicle path planning based on TLBO algorithm, International Journal on Smart Sensing & Intelligent Systems 7.3 (2014).
[16] Z. Zhai, G. Jia, and K. Wang., A novel teaching-learning-based optimization with error correction and cauchy distribution for path planning of unmanned air vehicle, Computational intelligence and neuroscience 2018 (2018) doi: 10.1155/2018/5671709
[17] Q. Niu, H. Zhang, and K. Li., An improved TLBO with elite strategy for parameters identification of pem fuel cell and solar cell models, International journal of hydrogen energy 39.8 (2014) 3837– 3854 doi: 10.1016/j.ijhydene.2013.12.110
[18] R. Rao, V. Savsani, and D. Vakharia., Teaching learning-based optimization: A novel method for constrained mechanical design optimization problems, Computer-Aided Design 43.3 (2011) 303 – 315 doi: 10.1016/j.cad.2010.12.015
[19] R. Rao and V. Patel., An elitist teaching-learning-based optimization algorithm for solving complex constrained optimization problems, International Journal of Industrial Engineering Computations 3.4 (2012) 535–560 doi: 10.5267/j.ijiec.2012.03.007